\documentclass{article}

\PassOptionsToPackage{square,numbers}{natbib}

\usepackage[final]{tackling_climate_workshop_style}

\usepackage[utf8]{inputenc} 
\usepackage[T1]{fontenc}    
\usepackage{url}            
\usepackage{booktabs}       
\usepackage{amsfonts}       
\usepackage{nicefrac}       
\usepackage{microtype}      
\usepackage{authblk}
\usepackage{multirow}
\usepackage{graphicx}
\usepackage{wrapfig,lipsum,booktabs}
\usepackage{caption, subcaption}
\usepackage[hidelinks]{hyperref}
\title{Deep Learning for Global Wildfire Forecasting}

\author[1,2]{Ioannis Prapas}
\author[1]{Akanksha Ahuja}
\author[1,2]{Spyros Kondylatos}
\author[1]{Ilektra Karasante}
\author[3]{Eleanna Panagiotou}
\author[4]{Lazaro Alonso}
\author[3]{Charalampos Davalas}
\author[3]{Dimitrios Michail}
\author[4]{Nuno Carvalhais}
\author[1]{Ioannis Papoutsis}

\affil[1]{Orion Lab, IAASARS\thanks{Institute for Astronomy, Astrophysics, Space Applications and Remote Sensing} , National Observatory of Athens}
\affil[2]{Image Processing Laboratory (IPL), Universitat de València}  
\affil[3]{Harokopio University of Athens}
\affil[4]{Max Planck Institute for Biogeochemistry}

\begin{document}

\maketitle

\begin{abstract}

Climate change is expected to aggravate wildfire activity through the exacerbation of fire weather.
Improving our capabilities to anticipate wildfires on a global scale is of uttermost importance for mitigating their negative effects.
In this work, we create a global fire dataset and demonstrate a prototype for predicting the presence of global burned areas on a sub-seasonal scale with the use of segmentation deep learning models.
Particularly, we present an open-access global analysis-ready datacube, which contains a variety of variables related to the seasonal and sub-seasonal fire drivers (climate, vegetation, oceanic indices, human-related variables), as well as the historical burned areas and wildfire emissions  for 2001-2021. 
We train a deep learning model, which treats global wildfire forecasting as an image segmentation task and skillfully predicts the presence of burned areas 8, 16, 32 and 64 days ahead of time. 
Our work motivates the use of deep learning for global burned area forecasting and paves the way towards improved anticipation of global wildfire patterns.

\end{abstract}

\section{Introduction}
Fire is a fundamental process in the Earth system, having an integral role in shaping the ecosystems around the globe. Although it has traditionally been considered as a net carbon-neutral process in the long-term, climate change can alter that with the exacerbation of fire weather, that exerts an upwards pressure on global fire activity \cite{jones_global_2022}. The expansion of fire to regions of evergreen forests could weaken their status as carbon sinks \cite{joshi_improving_2021, pausas_burning_2009}, releasing the stored carbon into the atmosphere, which in turn provides a feedback to climate change.
Additionally, wildfires pose a major natural hazard for societies globally causing loss of lives, properties, infrastructures and ecosystem services \cite{bowman_fire_2009, pettinari_fire_2020}. 
With changing climate and human activity increasingly affecting fire regimes around the world \cite{moritz_climate_2012, abatzoglou_impact_2016, jones_global_2022}, it is crucial to improve our understanding and anticipation capabilities of the wildfire phenomenon in the Earth system. 
This could foster the development of strategies to mitigate the negative effects of wildfires and climate change, with better forestry management, infrastructure resilience, disaster preparedness and development of more accurate warning systems. 

\citet{reichstein_deep_2019} and \citet{camps-valls_deep_2021} have argued to leverage the growing availability of open Earth Observation (EO) data 
and the advances in Machine Learning (ML) and particularly Deep Learning (DL) research to tackle Earth system problems. 
With respect to wildfire modeling, there is a limited amount of open-access datasets that combine variables related to the fire drivers and burned areas. Existing datasets are mostly focused on specific areas. 
For example, \citet{kondylatos_wildfire_2022} present a dataset for daily wildfire forecasting in the Eastern Mediterranean, while \citet{singla2021wildfiredb} and \citet{graff2021fireml} have published fire occurrence prediction datasets that cover the continental US. Moreover, \citet{huot_next_2022} have made available a fire spread dataset that also covers the US. 
The limited number of published datasets, as well as the challenges occurring when framing wildfire forecasting as an ML task \cite{prapas_deep_2021}, can be considered responsible for the limited amount of DL-related works dealing with the problem.
\citet{huot_deep_2020} use DL models that treat daily wildfire forecasting in the US as a segmentation task. \citet{kondylatos_wildfire_2022} use a variety of DL models to forecast fire danger in the Eastern Mediterranean,
\citet{yu_quantifying_2020} predict monthly fire in Africa with multi-layer perceptrons and highlight the contribution of oceanic indices. \citet{chen_forecasting_2020} predict global fire emissions on a sub-seasonal to seasonal scale and lastly \citet{zhang_deep_2021} predict global fire susceptibility on a monthly temporal resolution.

In this work, we present the \textbf{SeasFire Cube, a first of its kind curated global dataset (0.25$^{\circ}$ x 0.25$^{\circ}$ x 8-day resolution) for wildfire modeling for years 2001-2021 \cite{alonso_lazaro_2023}}; it combines historical burned areas and carbon emissions from wildfires, which can be used as target variables, along with variables related to the fire drivers (vegetation status, meteorological data, anthropogenic  factors, oceanic indices). To demonstrate a potential use of the dataset, we frame wildfire forecasting as an image segmentation task and train UNET++\cite{zhou_unet_2018} DL models to \textbf{predict the presence of global burned areas at different temporal horizons (8, 16, 32 and 64 days ahead)}. The models demonstrate forecasting skill that is evident both in the evaluation metrics and the visualisation of global prediction maps.

\section{Data and Methods}

\subsection{SeasFire Cube: A Datacube for Global Wildfire Modeling}\label{sec:data}

We gather global data related to the fire drivers (climate, vegetation, land cover, human presence), the burned areas from three different datasets and wildfire emissions for years 2001 to 2021, at a common spatio-temporal grid (0.25 $^{\circ}$ x 0.25 $^{\circ}$ x 8-day). 
We harmonize the data to create SeasFire Cube \cite{alonso_lazaro_2022_7108392}, an analysis-ready, open-access datacube stored in a cloud-friendly Zarr format \cite{zarr}, that can be equally used by Earth scientists and ML practitioners.

For a given timestep of the datacube, a variable carries information at a temporal resolution of the next 8 days. For example, a variable on January 1st, contains aggregated values from January 1st to January 8th. Additional details about the datacube are given in Appendix \ref{app:cube}. 

The SeasFire cube can be used for many downstream tasks related to wildfire modeling, such as forecasting burned area sizes or carbon emissions. In this study, we demonstrate how it can be used to forecast the burned area patterns 8 to 64 days ahead of time. 

For the modeling experiments that follow in the next sections, we choose a subset of the datacube variables, and one burned area product, which covers the range of 20 years from \citet{gwis_ds}, resampled to the datacube's grid. In total, 8 input variables are used based on their relevance and usage in burned area prediction tasks from existing literature \cite{jain_review_2020}: 2 EO variables from the MODIS satellite (day land surface temperature \cite{lst_ds} and the normalized difference vegetation index \cite{ndvi_ds}), 6 climatic variables from the ERA5 dataset \cite{era} (relative humidity, sea surface temperature, minimum temperature, total precipitation and vapour pressure deficit).

\begin{figure}[htbp]

  \centering
  \includegraphics[width=\linewidth]{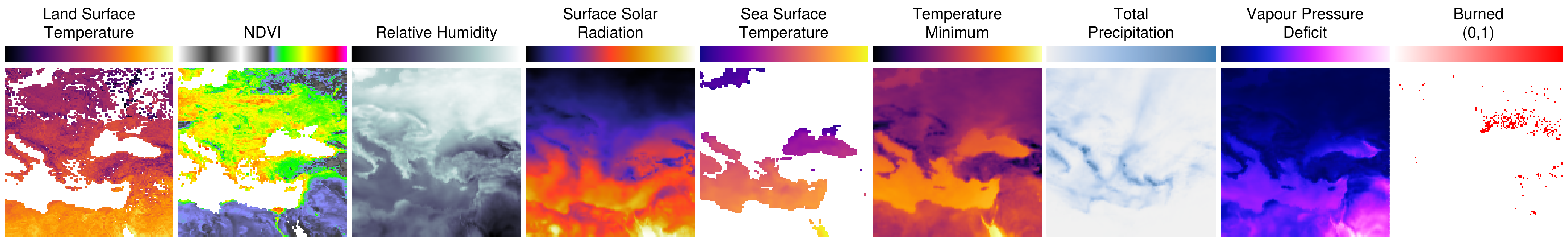}
  \caption{
    Visualization of the input and target variables.
  }
  \label{fig:variables}
  
\end{figure}

\subsection{Methods}\label{sec:methods}

\begin{figure}[htbp]
  \centering
  \includegraphics[width=1.0\linewidth]{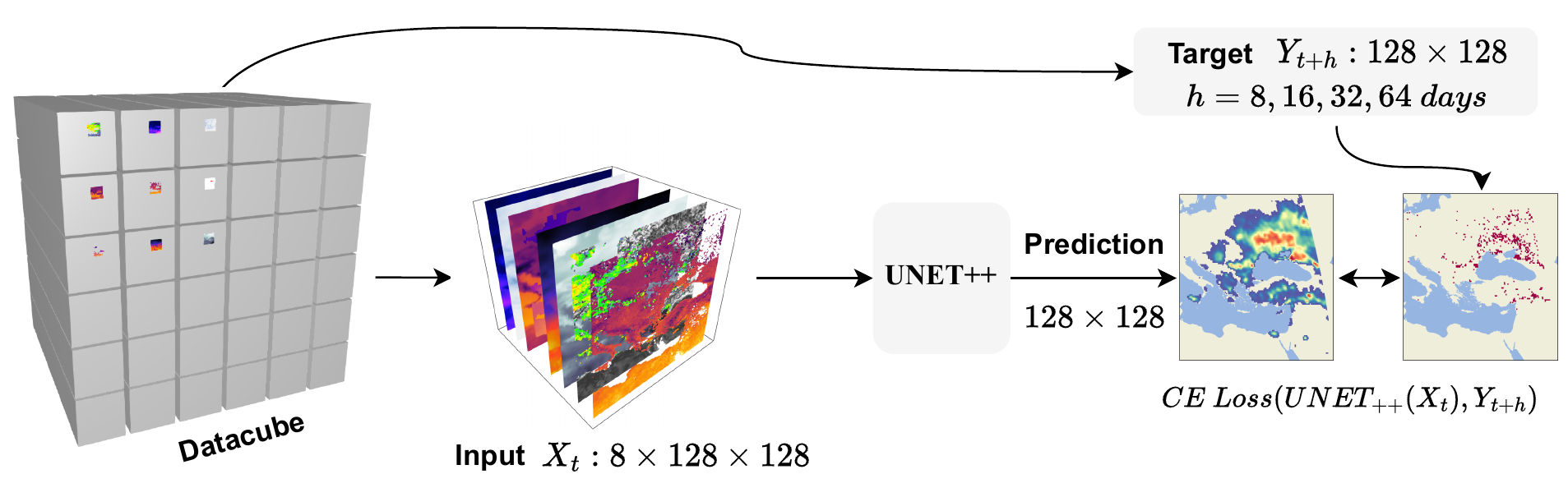}
  \caption{
    Pipeline diagram. The dataset is extracted from the datacube and fed to a UNET++ model that is trained with the Cross Entropy loss (CE loss) with inputs valid at time $t$ to predict the target (presence of burned areas) on time $t+h$. The prediction shown is an actual prediction from the validation set with $h=16$ days for an area covering the Eastern Mediterranean and its surroundings.
  }
  \label{fig:pipeline}
\end{figure}

From the SeasFire cube, we extract a subset of the features for our experiments and treat the burned area forecasting as a segmentation task where the input, target pairs have spatial dimensions 128 x 128 pixels, or 32 $^{\circ}$ x 32 $^{\circ}$. 
The 8 input variables, described previously (Section \ref{sec:data}), are stacked in a 8x128x128 tensor and fed as input to a UNET++ model \cite{zhou_unet_2018}. The target variable is chosen to be a burnt mask at a future time step. 
As such, the UNET is trained to predict the presence of burned areas at time $t+h$ using the input variables at time $t$. 
For each pixel, the burned area mask signifies the presence or absence of a burned area. This simplifies the ML task, by not taking into account the actual burned area size of the cell.  
A visualization of the input-target variables is shown in Figure \ref{fig:variables} and the ML pipeline in Figure \ref{fig:pipeline}.

 To account for the imbalance between the negative (not burned) and the positive (burned) class, we train on patches that include at least one burned area pixel. This discards about half of the examples, mainly on polar regions or in the sea, ending up with 22,743 out of the 24,012 patches. 
 Dealing with a forecasting task, we do a time split using years from 2002 to 2017 for training (20,178  examples), 2018 for validation (1,261 examples) and 2019 for testing (1,304 examples). Among the total pixels of those datasets, only about 1.6\% are of the positive class (burned). The split is based on the burned area date, which means that the resulting datasets for the different forecasting horizons have the same set of targets, even if the input varies.

 The forecasting lead time $h$ is in multiples of 8 days, which is the temporal resolution of the dataset. 
 A different model is trained for each of the four different forecast lead times ($h$ = 8, 16, 32, 64 days). As evaluation metrics, we use the Area Under the Precision-Recall Curve (AUPRC) and F1-score, which are appropriate for imbalanced datasets. Area Under the Receiver Operator Characteristic (AUROC) is also reported to demonstrate that it is not a good fit for the specific problem. 
 
Weekly Mean Seasonal Cycle baseline: We calculate the temporal average of burned area sizes for each one of the weeks in a year, on the training years (2002 - 2018). Each patch on the validation and test set is compared to an equal dimension patch that contains the average of the respective week of the historical dataset. The AUPRC and the AUROC of the weekly averages are also reported. 
 
We use the EfficientNet-b1 encoder \cite{tan2019efficientnet}, and train for 30 epochs with the cross entropy loss, which is commonly used for segmentation tasks. 
The model with the best validation loss across the epochs is kept for testing.
Each experiment takes around 1.25 GPU hours on an NVIDIA RTX 3090 (24GB).

\section{Results}

On the test set, we observe that the models achieve an AUPRC larger than 0.55 for all forecasting lead times (Table \ref{tab:metrics}). 
Significantly, the models demonstrate a higher predictive skill than the weekly mean seasonal cycle, which shows that they have learned more than merely the seasonality of burned area patterns. The F1-score and the AUPRC are larger when the forecasting window is shorter. The high AUROC, reaching 0.931 for the baseline, inflates the performance on such an imbalanced datasets.

\begin{figure}[htbp]
  \centering
  \includegraphics[width=\linewidth]{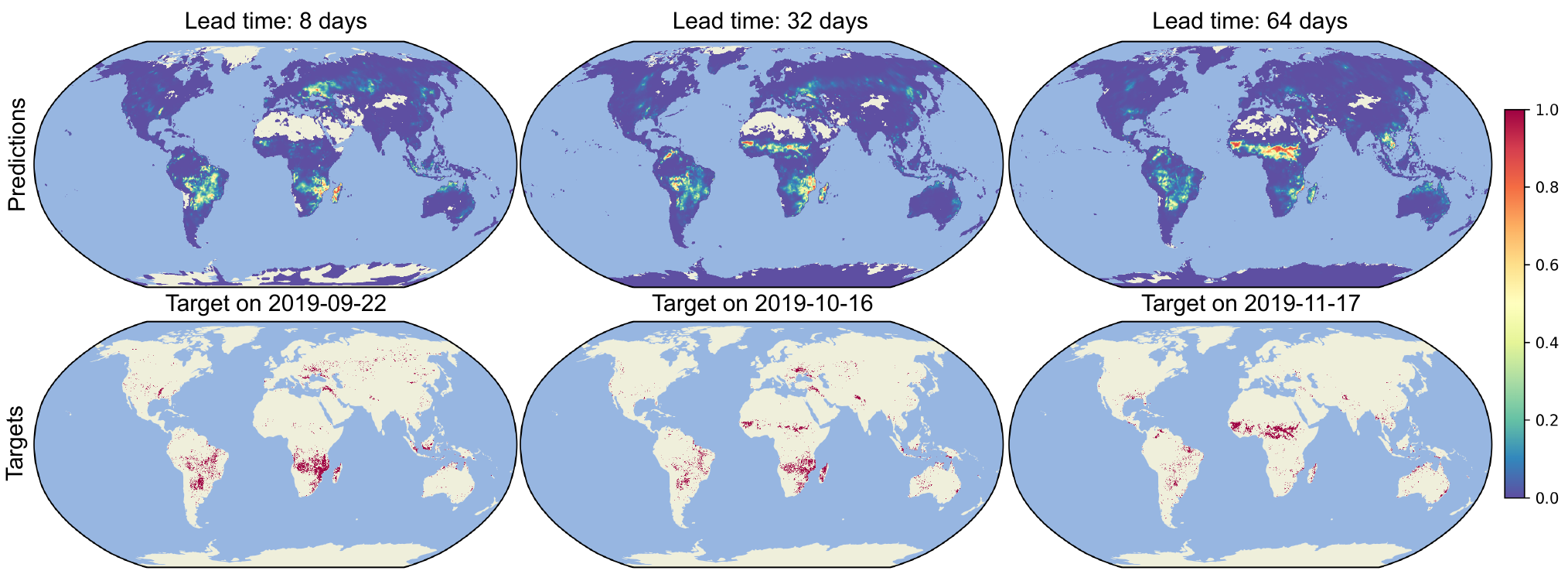}
   
  \caption{
    Prediction (top row) and target (bottom row) maps for lead forecasting times of 8 (left), 32 (middle) and 64 (right) days. Predictions lower than $0.0001$ are visualized as missing values. The input for all predictions is the same from 2019-09-14.
  }
  \label{fig:maps}

\end{figure}

The global predictions in the test year and the equivalent target maps (Figure \ref{fig:maps}) illustrate that the DL models capture several large-scale burned area patterns. Notably, the models capture the change of fire activity in eastern Europe and south-east Asia, as well as the shift in fire pattern from the southern to the northern African savanna. Although burned area size is not explicitly modelled, Africa, which accounts for around 70\% of the global burned areas \cite{yu_quantifying_2020}, seems to dominate the focus of the model. It is not clear whether regional patterns are captured and it needs further future investigation. Moreover, we see that the DL models have learned some recognisable patterns, for example the sea, the poles and the Sahara desert are always predicted to be in low danger zones.

\begin{table}[htbp]
\caption{AUPRC, F1-score for the UNET++ model forecasting with different lead times on the test dataset (year 2019). Baseline values for the weekly mean seasonal cycle also reported.}
\label{tab:metrics}
\centering
\begin{tabular}{ccccc}
\toprule
                                 & Lead time (days) & AUPRC & F1-score & AUROC\\ 
\midrule
\multirow{4}{*}{\textbf{UNET++}} & 8                & 0.5891 & 0.5271 & 0.9754 \\
                                 & 16               & 0.5851 & 0.5231 & 0.9756 \\
                                 & 32               & 0.5734 & 0.5167 & 0.9781 \\
                                 & 64               & 0.5598 & 0.4947 & 0.9782 \\
\midrule

Weekly Mean Seasonal Cycle & - & 0.4519 & - & 0.9306 \\

\bottomrule
\end{tabular}
\end{table}

\section{Discussion}

Our work demonstrates that segmentation DL models can skilfully predict global burned area patterns from a snapshot of fire driver input data valid from 8 days to more than 2 months in advance. This does not imply that this is the best architecture or setup for the task, but rather motivates the use of DL for burned area forecasting and calls for future work that is facilitated by our curated analysis-ready datacube \cite{alonso_lazaro_2022_7108392}.
Future work will enhance the pipeline to take into account time-series of the input variables. Furthermore, we can use more predictors and measure their contribution. 
Rather than predicting the presence of burned area, it could be more insightful to predict the actual burned area size. 
Moreover, an extensive evaluation is appropriate; we want to quantify to what degree the DL models capture the temporal variability of fire, including mega fires due to climate extremes, on sub-seasonal, seasonal and annual scales and if they can also capture regional patterns in diverse geographical locations. We can use metrics considering the temporal variability of burned areas \cite{joshi_improving_2021}. 

With this demonstration on the SeasFire cube, we hope to spark the interest of the research community for global wildfire modeling, an immensely complex and important task of our time.
The datacube includes a unique set of variables that will allow for a number of different modeling tasks. 
Indicatively, time-series of oceanic indices allow to model spatio-temporal teleconnections. 
Human-related variables such as population density and land cover can provide insights in the effects of humans on wildfire at different temporal scales. 
Carbon emissions from wildfires are also part of the dataset and can be used as to study the impact of wildfires on the global carbon cycle.

\section{Conclusion}

In the context of climate change, we believe it is important to move from measuring regional effects in isolation to treating the Earth as one interconnected system. We present a versatile global dataset that includes multiple variables to capture the multiple facets of the wildfires. The open-access dataset is available to the community to analyse and model global wildfires from 2001 to 2021. As a demonstration, we have framed the forecasting of burned areas as a segmentation task and showed how DL models are skilful at forecasting global burned area patterns even 2 months in advance.

\begin{ack}
This work is part of the SeasFire project, which deals with "Earth System Deep Learning for Seasonal Fire Forecasting" and is funded by the European Space Agency (ESA)  in the context of ESA Future EO-1 Science for Society Call.
\end{ack}

\bibliography{references.bib}

\newpage

\appendix
\section{SeasFire Cube: A Global Dataset for Seasonal Fire Modeling}

The SeasFire Cube \cite{alonso_lazaro_2022_7108392} combines a variety of datasets at a common spatiotemporal grid to ease the analysis of global wildfire activity. Models developed on the dataset may not be of direct operational use due to the fact that the ERA5 reanalysis dataset is not available in real-time. Due to that, the dataset is valuable to compare different approaches, but may overestimate their performance. For use in production, it is suggested to switch to from ERA5 to archived forecasts.

\label{app:cube}
\begin{table}[h]
\resizebox{\textwidth}{!}{%
\begin{tabular}{llll}
\hline
\textbf{Full   name}                   & \textbf{Units}  & \textbf{Aggregation} & \textbf{Comment}                               \\ \hline
\textit{\textbf{Dataset:   Copernicus ERA5} \cite{era} }       &                               &                 &                      \\ \hline
Mean sea level pressure                                     & Pa                            & 8-day mean & -                    \\
Total precipitation                                         & m                             & 8-day sum  & -                    \\
Relative humidity                      & \%          & 8-day mean      & in-house calculation                           \\
Vapor Pressure Deficit                                      & hPa                           & 8-day mean & in-house calculation \\
Sea Surface Temperature                                     & K                             & 8-day mean & -                    \\
Skin temperature                                            & K                             & 8-day mean & -                    \\
Wind speed at 10 meters                                     & m s-2                         & 8-day mean & in-house calculation \\
Temperature at 2 meters - Mean                              & K                             & 8-day mean & -                    \\
Temperature at 2 meters - Min                               & K                             & 8-day min   & -                    \\
Temperature at 2 meters -   Max                             & K                             & 8-day max  & -                    \\
Surface net solar radiation                                 & MJ m-2                        & 8-day mean & -                    \\
Surface solar radiation downwards                           & MJ m-2                        & 8-day mean & -                    \\
Volumetric soil water level 1                               & unitless                      & 8-day mean & -                    \\ \hline
\textit{\textbf{Dataset: Copernicus CEMS } \cite{cems_ds} }          &                               &                 &                      \\ \hline
Drought Code Maximum                                        & unitless                      & 8-day max  & -                    \\
Fire Weather Index Maximum                                  & unitless                      & 8-day max  & -                    \\
Fire Weather Index Average                                  & unitless                      & 8-day mean & -                    \\ \hline
\textit{\textbf{Dataset: Copernicus CAMS } \cite{cams_ds} }          &                               &                 &                      \\ \hline
Carbon dioxide emissions from wildfires                     & m-2 kg s-1                    & 8 day sum  & -                    \\
Fire radiative power                                        & W m-2                         & 8-day sum  & -                    \\ \hline
\textit{\textbf{Dataset: Copernicus FCCI } \cite{fcci_ds} }          &                               &                 &                      \\ \hline
Burned Areas                                                & ha                            & 8-day sum  & -                    \\ \hline
\textit{\textbf{Dataset: Nasa MODIS land products}} &                               &                 &                      \\ \hline
Land Surface temperature at day                             & K                             & 8-day mean & resampled at 0.25 $^{\circ}$x 0.25 $^{\circ}$(MOD11C1 v006 \cite{lst_ds})        \\
Leaf Area Index                                             & unitless                      & 8-day mean & resampled at 0.25 $^{\circ}$x 0.25 $^{\circ}$(MCD15A2H v006  \cite{lai_ds})      \\
Normalized Difference Vegetation Index & unitless        & 8-day mean & resampled at 0.25 $^{\circ}$x 0.25 $^{\circ}$(MOD13C1 v006  \cite{ndvi_ds})                                 \\ \hline
\textit{\textbf{Dataset: Nasa SEDAC } \cite{pop_ds} }               &                               &                 &                      \\ \hline
Population density                                          & Persons per square kilometers &  -              & -                    \\ \hline
\textit{\textbf{Dataset: GFED } \cite{gfed_ds} }                     &                               &                 &                      \\ \hline
Burned Areas (large fires only)                             & ha                            & 8-day sum  & -                    \\ \hline
\textit{\textbf{Dataset: GWIS } \cite{gwis_ds} }                     &                               &                 &                      \\ \hline
Burned Areas                           & ha           & 8-day sum       & Rasterized and resampled at 0.25 $^{\circ}$x 0.25 $^{\circ}$\\ \hline
\textit{\textbf{Dataset: NOAA  } \cite{noaa_ds} }                    &                               &                 &                      \\ \hline
Western Pacific Index                                       & unitless                      & -               & -                    \\
Pacific North American Index                                & unitless                      & -               & -                    \\
North Atlantic Oscillation                                  & unitless                      & -               & -                    \\
Southern Oscillation Index                                  & unitless                      & -               & -                    \\
Global Mean Land/Ocean Temperature                          & unitless                      & -               & -                    \\
Pacific Decadal Oscillation                                 & unitless                      & -               & -                    \\
Eastern Asia/Western Russia                                 & unitless                      & -               & -                    \\
East Pacific/North Pacific Oscillation                      & unitless                      & -               & -                    \\
Nino 3.4 Anomaly                                            & unitless                      & -               & -                    \\
Pacific North American Index                                & unitless                      & -               & -                   
\end{tabular}%
}
\end{table}
\end{document}